\documentclass[runningheads]{llncs}
\usepackage[T1]{fontenc} 
\usepackage{graphicx}
\usepackage{amsmath, amssymb}
\usepackage{hyperref}
\usepackage{tikz} 
\usepackage{braket} 
\usepackage{bm} 
\usepackage{bbding} 

\definecolor{vnecolor}{RGB}{30, 144, 255}

\newcommand{\R}{\mathbb{R}}
\newcommand{\C}{\mathbb{C}}

\newcommand{\V}{\mathcal{V}} 
\newcommand{\Tr}{\operatorname{Tr}}
\newcommand{\Id}{\mathbb{I}} 
\newcommand{\vect}{\operatorname{vec}} 

\newcommand{\Hp}{\mathcal{H}} 
\newcommand{\Lop}{\mathcal{L}} 
\newcommand{\D}{\mathcal{D}}   
\newcommand{\E}{\mathcal{E}}   

\begin{document}

\title{Density Matrix RNN (DM-RNN): A Quantum Information Theoretic Framework for Modeling Musical Context and Polyphony}

\titlerunning{Density Matrix RNN (DM-RNN)}

\author{Joonwon Seo\inst{1}\orcidID{0009-0000-5162-9915} \and
Mariana Montiel\inst{1}\orcidID{0000-0002-7859-1547}\Envelope}

\authorrunning{J. Seo and M. Montiel}

\institute{Department of Mathematics and Statistics,\\
Georgia State University,\\
Atlanta, GA 30303, USA\\
\email{\{jseo27, mmontiel\}@gsu.edu}}

\maketitle

\begin{abstract}
Classical Recurrent Neural Networks (RNNs) summarize musical context into a deterministic hidden state vector $h_t \in \R^d$, imposing an information bottleneck that fails to capture the inherent ambiguity in music.
We propose the Density Matrix RNN (DM-RNN), a novel theoretical architecture utilizing the Density Matrix $\rho_t \in \D(\Hp_d)$.
This allows the model to maintain a statistical ensemble of musical interpretations (a mixed state), capturing both classical probabilities and quantum coherences.
We rigorously define the temporal dynamics using Quantum Channels (CPTP maps).
Crucially, we detail a parameterization strategy based on the Choi-Jamio\l{}kowski isomorphism, ensuring the learned dynamics remain physically valid (CPTP) by construction.
We introduce an analytical framework using Von Neumann Entropy $S(\rho)$ to quantify musical uncertainty and Quantum Mutual Information (QMI) to measure entanglement between voices.
The DM-RNN provides a mathematically rigorous framework for modeling complex, ambiguous musical structures.
\keywords{Computational Music Theory \and Density Matrix \and Quantum Information Theory \and CPTP Maps \and Choi Matrix \and Recurrent Neural Networks \and Von Neumann Entropy.}
\end{abstract}

\section{Introduction}

The computational modeling of musical sequences presents a profound challenge. Music is characterized by intricate temporal dependencies and inherent ambiguity~\cite{Huron2006}. At any given time step $t \in \mathbb{Z}^+$, a musical piece often implies not a single continuation, but a distribution over possibilities.

Traditional sequence models, such as LSTM~\cite{Hochreiter1997} or Transformers~\cite{Vaswani2017}, compress the preceding context into a single hidden state vector, $h_t \in \mathbb{R}^{d}$. This vector representation forces the model to commit to a point estimate, collapsing the rich tapestry of musical possibilities and struggling to capture nuanced ambiguities.

We propose a novel theoretical framework, the Density Matrix RNN (DM-RNN), inspired by the mathematical formalism of quantum mechanics, which is inherently suited to describing systems characterized by statistical uncertainty and interference.

\subsection{The Density Matrix Approach}

The cornerstone of our approach is the Density Matrix, $\rho$. Operating on a complex Hilbert space $\Hp$, $\rho$ provides a complete description of the statistical state of a system. By employing $\rho$ as the hidden state, we model the musical context not as a definitive state, but as a \textit{statistical ensemble} of possibilities (a mixed state).

The DM-RNN replaces the traditional vector $h_t \in \R^d$ with a density matrix $\rho_t \in \D(\Hp_d)$. This formalism offers a richer representation:

\begin{enumerate}
    \item \textbf{Populations (Diagonal Elements):} Represent the classical probabilities of the system being in specific latent basis states.
    \item \textbf{Coherences (Off-Diagonal Elements):} Encode the stable phase relationships between these states, mathematically capturing how different musical interpretations interfere or are structurally linked.
\end{enumerate}

\subsection{Rigorous Temporal Dynamics and Parameterization}

We formalize the temporal evolution, $\rho_{t-1} \mapsto \rho_t$, using Quantum Channels, defined as Completely Positive Trace-Preserving (CPTP) maps. Ensuring that a learned neural network transformation adheres to these constraints is non-trivial. A key contribution of this work is the rigorous definition of the model's parameterization via the Choi-Jamio\l{}kowski isomorphism, which allows us to enforce the CPTP constraints through a differentiable normalization scheme applied to the Kraus operators (Section 4.4).

\subsection{A New Analytical Framework}

The DM-RNN framework enables analysis grounded in Matrix Theory~\cite{Horn2013} and quantum information metrics. We utilize the Von Neumann Entropy (VNE), $S(\rho)$, to quantify musical uncertainty. Spectral Analysis of $\rho_t$ reveals the dominant musical eigenstates. We address polyphony by introducing Quantum Mutual Information (QMI) to quantify entanglement, provided a specific tensor product structure is imposed (Section 5.3).

\subsection{Contributions}

The main contributions of this work are theoretical:

\begin{enumerate}
    \item The proposal of the DM-RNN architecture.
    \item A rigorous mathematical definition of its parameterization using the Choi matrix to guarantee CPTP dynamics.
    \item A novel analytical framework for computational musicology based on quantum information metrics (VNE, QMI) and Matrix Theory.
\end{enumerate}

This paper focuses on the theoretical development. Practical implementation involves significant computational challenges, suggesting the need for efficient methods like Tensor Networks, discussed in Section 6.3.

\section{Background and Related Work}

We first rigorously examine the mathematical framework of classical RNNs and contextualize the DM-RNN within existing approaches.

\subsection{Sequence Modeling Formalism}

In the standard sequence modeling formalism, a finite vocabulary $\mathcal{V}$ is defined. A sequence is denoted as $X=(x_1, \dots, x_T)$, where $x_t \in \mathcal{V}$.
The objective is to estimate $P(X)$, decomposed autoregressively. When $x_{<t}$ denotes the predecessor sequence $(x_1, \dots, x_{t-1})$, the probability $P(X)$ is given by:

\begin{equation}
P(X) = \prod_{t=1}^{T} P(x_t | x_{<t})
\label{eq:chain_rule}
\end{equation}
where the condition for $t=1$ (i.e., $x_{<1}$) is empty.

\subsection{The Classical RNN Architecture}

Classical RNNs maintain a hidden state $h_t \in \mathbb{R}^d$. Inputs are mapped to embeddings $e_t \in \R^m$. The evolution is governed by a parameterized transition function $f: \mathbb{R}^d \times \mathbb{R}^m \rightarrow \mathbb{R}^d$:

\begin{equation}
h_t = f(h_{t-1}, e_t; \theta)
\label{eq:rnn_update}
\end{equation}
The output distribution is given by the softmax function $\sigma$:

\begin{equation}
P(x_{t+1} | x_{<t}; \theta) = \sigma(W_{\text{out}}h_t + b_{\text{out}})
\label{eq:rnn_output}
\end{equation}
where $W_{\text{out}} \in \mathbb{R}^{|\mathcal{V}| \times d}$ and $b_{\text{out}} \in \mathbb{R}^{|\mathcal{V}|}$.

\subsection{The Representational Bottleneck}

The critical limitation of the classical RNN approach is that the hidden state $h_t$ is a \textit{point estimate} in $\R^d$. This creates a bottleneck when modeling ambiguity. If we consider two distinct musical interpretations $h_A, h_B \in \R^d$, the RNN might learn an interpolation, $h_t \approx \alpha h_A + (1-\alpha)h_B$, with $\alpha \in [0, 1]$. This interpolation is fundamentally different from a statistical ensemble, potentially leading to information loss.

\subsection{Classical Approaches to Uncertainty}

Bayesian RNNs~\cite{Fortunato2017} and Variational RNNs (VRNNs)~\cite{Chung2015} introduce stochasticity. These methods primarily focus on capturing \textit{epistemic uncertainty} (about parameters) or \textit{aleatoric uncertainty} (data variability). They typically model the hidden state as a classical probability distribution over the vector space $\R^d$.

\subsection{Quantum-Inspired Models and Tensor Networks}

The application of quantum formalisms to machine learning is an emerging field~\cite{Biamonte2017}. The density matrix formalism has been explored in various domains. In Natural Language Processing (NLP), density matrices have been used to represent word embeddings, capturing polysemy and hyponymy as mixed states~\cite{Bankova2019}.

\textbf{Quantum Sonification.} In the auditory domain, the intersection with quantum formalisms is also an active area of research. Recent work by Christie and Trayford~\cite{Christie2024} introduced a framework for "Open Quantum Sonification," mapping density matrix elements and their phases to auditory signals. Specifically, they render off-diagonal coherences as binaural signals to intuitively illustrate quantum phenomena like decoherence. While these approaches focus on analyzing physical quantum systems via sound, the DM-RNN conversely employs the density matrix formalism and CPTP dynamics as a rigorous generative framework to model the implicit ambiguity and coherence within musical structures themselves.

Relatedly, Tensor Networks (TN), originating from condensed matter physics for efficiently representing high-dimensional quantum states~\cite{Orus2014}, have also been applied to classical machine learning, including sequence modeling~\cite{Stoudenmire2016}. TN approaches often utilize Matrix Product States (MPS) to model sequences, focusing on capturing correlations efficiently.

DM-RNN is distinct in its approach. While related to TN models~\cite{Stoudenmire2016}, DM-RNN explicitly enforces quantum-valid dynamics (CPTP maps) for state evolution, rather than optimizing a classical MPS representation. Compared to attention-based models like the Music Transformer~\cite{Huang2019}, which compute long-range dependencies via attention scores, DM-RNN focuses on the rigorous evolution of the state representation itself ($\rho_t$) to inherently capture ambiguity and coherence. The density matrix offers a richer representation than classical probability distributions by explicitly capturing coherences (off-diagonal elements), representing phase relationships between different interpretations (Section 5.1.2).

\section{The Density Matrix Formalism in Music}

We adopt the Density Matrix formalism to represent musical context.

\subsection{Mathematical Preliminaries and Definitions}

We define the state space within a $d$-dimensional complex Hilbert space, $\mathcal{H}_d \cong \mathbb{C}^d$. The use of a complex space, rather than a real space $\R^d$, is essential as the off-diagonal elements (coherences) require complex numbers to encode the phase relationships between basis states, which are necessary to distinguish between classical mixtures and coherent superpositions (Section 5.1.2). Let $\Lop(\Hp_d)$ denote the space of linear operators on $\Hp_d$.

A pure state is a normalized vector $\ket{\psi} \in \mathcal{H}_d$. The corresponding density matrix is $\rho_{\text{pure}} = \ket{\psi}\bra{\psi}$.

\textbf{Mixed States.} A \textit{mixed state} represents a statistical ensemble $\{(p_i, \ket{\psi_i})\}_{i=1}^{N}$, where $p_i \ge 0$ and $\sum_i p_i = 1$. The Density Matrix $\rho \in \Lop(\Hp_d)$ is the convex combination:

\begin{equation}
\rho = \sum_{i=1}^{N} p_i \ket{\psi_i}\bra{\psi_i}
\label{eq:density_matrix_def}
\end{equation}

\textbf{The Set of Density Matrices.} The set of all valid density matrices $\D(\Hp_d)$ is defined as:
\begin{equation}
\D(\Hp_d) := \{ \rho \in \Lop(\Hp_d) \mid \rho = \rho^\dagger, \Tr(\rho) = 1, \rho \ge 0 \}
\end{equation}
This is a convex set defined by Hermiticity, Unit Trace, and Positive Semidefiniteness (PSD).

\subsection{Musical Interpretation and the Role of the Basis}

When $\rho_t \in \D(\Hp_d)$ is the hidden state, its interpretation relies on an orthonormal basis $\{\ket{u_i}\}_{i=1}^d$ for $\mathcal{H}_d$. In a neural network, this basis emerges during training. While the specific interpretation of populations ($\rho_{ii}$) and coherences ($\rho_{ij}$) is basis-dependent, key metrics (such as entropy) are invariant under unitary changes of basis.

\textbf{Basis Stabilization during Training.}
Although the density matrix formalism possesses unitary gauge freedom, the DM-RNN architecture stabilizes the interpretation basis through the learned POVM operators $\{M_v\}$ (Section 4.5). The optimization objective (e.g., negative log-likelihood based on the Born rule) forces the model to utilize a specific basis that optimally maps the internal state $\rho_t$ to the output vocabulary $\V$, ensuring a consistent internal model of the musical structure.

\subsection{Quantifying Musical Ambiguity}

The density matrix provides intrinsic, basis-independent metrics for quantifying ambiguity.

\textbf{Purity.} The purity $\gamma(\rho) := \Tr(\rho^2)$ is bounded by $1/d \le \gamma(\rho) \le 1$. $\gamma(\rho)=1$ iff the state is pure.

\textbf{Von Neumann Entropy (VNE).} The most comprehensive measure of mixedness is the Von Neumann Entropy, $S: \D(\Hp_d) \rightarrow \R_{\ge 0}$:

\begin{equation}
S(\rho) := -\Tr(\rho \log_2 \rho) = -\sum_{i=1}^{d} \lambda_i \log_2 \lambda_i
\label{eq:vne}
\end{equation}
where $\{\lambda_i\}$ are the eigenvalues of $\rho$. $S(\rho)$ quantifies the uncertainty inherent in the mixed state (in bits).

\section{Proposed Architecture: The DM-RNN}

We introduce the DM-RNN architecture, focusing on the rigorous mathematical construction of its temporal evolution and parameterization.

\subsection{The Architecture Overview}

The DM-RNN processes an input embedding $e_t \in \mathbb{R}^m$ and updates its hidden state $\rho_{t-1} \in \D(\Hp_d)$. The recurrence relation is defined by a parameterized transition function $\mathcal{F}: \D(\Hp_d) \times \mathbb{R}^m \rightarrow \D(\Hp_d)$:

\begin{equation}
\rho_t = \mathcal{F}(\rho_{t-1}, e_t; \theta)
\label{eq:dmrnn_update}
\end{equation}

\subsection{Temporal Dynamics via Quantum Channels (CPTP Maps)}

The transition function $\mathcal{F}$ must be implemented as a Quantum Channel, $\E: \Lop(\Hp_d) \rightarrow \Lop(\Hp_d)$, which must be Completely Positive and Trace-Preserving (CPTP)~\cite{Nielsen2010}.

\textbf{The Necessity of CPTP Maps.}
\begin{itemize}
    \item \textbf{Trace-Preserving (TP):} Ensures $\Tr(\E(\rho)) = 1$.
\item \textbf{Completely Positive (CP):} Ensures that for any auxiliary Hilbert space $\Hp_A$, the induced map $(\E \otimes \Id_A)$ (where $\otimes$ denotes the tensor product and $\Id_A$ is the identity map on $\Hp_A$) remains positive, guaranteeing physical validity.
\end{itemize}

\subsection{Kraus Operator Representation}

A CPTP map $\E$ can always be represented by the Kraus decomposition:

\begin{equation}
\E(\rho) = \sum_{k=1}^{K} K_k \rho K_k^\dagger
\label{eq:kraus_update}
\end{equation}
where $\{K_k\} \subset \Lop(\Hp_d)$ are Kraus Operators ($K \le d^2$) satisfying the completeness relation:

\begin{equation}
\sum_{k=1}^{K} K_k^\dagger K_k = \Id_d
\label{eq:kraus_completeness}
\end{equation}

\subsection{Rigorous Parameterization of the CPTP Map}

The core challenge is defining a parameterized function that generates the channel $\E_t(e_t; \theta)$ while strictly enforcing the CPTP constraints. We achieve this by utilizing the Choi-Jamio\l{}kowski isomorphism to ensure Complete Positivity (CP) and a differentiable normalization scheme to ensure Trace Preservation (TP).

\textbf{The Choi-Jamio\l{}kowski Isomorphism.} There is a one-to-one correspondence between linear maps $\E$ and Choi matrices $C_\E \in \Lop(\Hp_d \otimes \Hp_d)$. The isomorphism is established by applying the channel $\E$ to one part of a maximally entangled state. Specifically, $C_\E := (\E \otimes \Id)(\ket{\Phi^+}\bra{\Phi^+})$, where $\ket{\Phi^+} = \sum_{i=1}^d \ket{u_i}\otimes\ket{u_i}$ is the (unnormalized) maximally entangled state and $\Id$ is the identity map. $\E$ is CP iff $C_\E \ge 0$. $\E$ is TP iff the partial trace over the output space satisfies $\Tr_{\text{out}}(C_\E) = \Id_d$.

\textbf{Parameterization Construction.}
We construct the parameterization by generating and normalizing the Kraus operators directly, ensuring CP and TP by construction in a differentiable manner.

\begin{enumerate}
    \item \textbf{Enforcing CP via Cholesky Factor:} We utilize the property that the columns of the Cholesky factor of the Choi matrix correspond to the vectorized Kraus operators. Let $G$ be a neural network parameterized by $\theta_G \subset \theta$, which generates a complex matrix $L_t \in \C^{d^2 \times d^2}$ (the Cholesky factor):
    \begin{equation}
    L_t = G(e_t; \theta_G)
    \end{equation}
    The corresponding Choi matrix $C'_{\E_t} = L_t L_t^\dagger$ is PSD by construction, guaranteeing CP.

    \item \textbf{Extracting Kraus Operators:} We obtain a set of unnormalized Kraus operators $\{K'_k\}_{k=1}^{d^2}$ by reshaping (unvectorizing) the columns of $L_t$:
    \begin{equation}
    K'_k(t) = \operatorname{mat}(L_t[:, k])
    \end{equation}
    where $\operatorname{mat}(\cdot)$ is the inverse vectorization operation ($\vect^{-1}$), reshaping the $d^2 \times 1$ column vector into a $d \times d$ matrix.

    \item \textbf{Enforcing TP via Differentiable Normalization:} We enforce the completeness relation (Eq.~\ref{eq:kraus_completeness}) via a normalization layer~\cite{Trouvain2020}. First, we compute the normalization matrix $S_t \in \C^{d\times d}$:
    \begin{equation}
    S_t = \sum_{k=1}^{d^2} (K'_k(t))^\dagger K'_k(t)
    \end{equation}
    We normalize the operators:
    \begin{equation}
    K_k(t) = K'_k(t) S_t^{-1/2}
    \label{eq:kraus_normalization}
    \end{equation}
    To ensure $S_t$ is invertible and maintain numerical stability, we use a regularized inverse square root (e.g., by adding a small $\epsilon \Id_d$ to $S_t$, where typically $\epsilon \approx 10^{-6}$). This operation is differentiable, enabling end-to-end training.
\end{enumerate}

\textbf{The DM-RNN Update.} The state update is performed using the normalized Kraus operators via Eq.~\ref{eq:kraus_update}. This approach rigorously guarantees that $\E_t$ is CPTP and avoids the computationally expensive eigen-decomposition of the full Choi matrix (which scales as $O(d^6)$).
\subsection{Prediction via Measurement (POVM)}

To predict $x_{t+1} \in \V$, we use Positive Operator-Valued Measures (POVMs). We define a set of learned measurement operators $\{M_v\}_{v \in \V} \subset \Lop(\Hp_d)$, satisfying Positivity ($M_v \ge 0$) and Completeness ($\sum_{v\in \V} M_v = \Id_d$).

The probability of observing event $v$ is given by the Born rule:

\begin{equation}
P(x_{t+1}=v|\rho_t) = \Tr(M_v \rho_t)
\label{eq:born_rule}
\end{equation}

\textbf{POVM Parameterization.} We introduce auxiliary learned matrices $\{A_v\}_{v \in \V} \subset \C^{d\times d}$, parameterized by $\theta_M \subset \theta$.
We define $M'_v = A_v^\dagger A_v$ (guaranteeing positivity). We enforce completeness by normalization: Let $S = \sum_{v\in\V} M'_v$. The valid POVM elements are:
\begin{equation}
M_v = S^{-1/2} M'_v S^{-1/2}
\end{equation}
To handle potential singularity of $S$ and ensure numerical stability, we utilize a regularized inverse square root (e.g., using $S + \epsilon \Id_d$ for a small $\epsilon > 0$, or the Moore-Penrose pseudo-inverse), similar to the normalization strategy employed in Section 4.4.

\section{Matrix Theory and Musical Analysis}

The DM-RNN framework establishes a rigorous foundation for analysis grounded in Matrix Theory and Quantum Information Theory.

\subsection{Quantifying Musical Uncertainty and Coherence}

The ambiguity within the context $\rho_t$ is measured by the Von Neumann Entropy $S(\rho_t)$ (Eq.~\ref{eq:vne}).

\begin{equation}
S(\rho_t) = -\sum_{i=1}^d \lambda_i(t) \log_2 \lambda_i(t)
\label{eq:vne_t}
\end{equation}
The evolution of $S(\rho_t)$ tracks musical tension (mixing) and resolution (purification).

\subsubsection{Illustration: Modeling Ambiguity via Mixed States}

We illustrate with an example in $\Hp_2 \cong \C^2$, spanned by $\ket{A}$ (Tonic) and $\ket{B}$ (Dominant).

\begin{enumerate}
    \item \textbf{t=0:} Clear A. $\rho_0 = \ket{A}\bra{A}$. $S(\rho_0) = 0$.
    \item \textbf{t=1 (Tension):} A and B are equally plausible.
    \begin{itemize}
        \item DM-RNN: Maximally mixed state: $\rho_1 = \begin{pmatrix} 0.5 & 0 \\ 0 & 0.5 \end{pmatrix}$. $S(\rho_1) = 1$ bit.
        \item Classical RNN: Interpolation: $h_1 \approx 0.5 h_A + 0.5 h_B$. Potential information loss.
    \end{itemize}
    \item \textbf{t=2:} Resolved to B. $\rho_2 = \ket{B}\bra{B}$. $S(\rho_2) = 0$.
\end{enumerate}

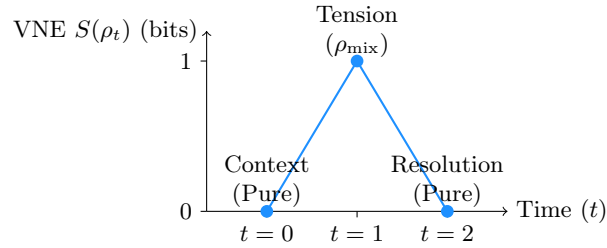
\begin{figure}[t]
\centering
\begin{tikzpicture}[scale=0.8]
    \def\yscale{2.5}

    \draw[->] (0,0) -- (5,0) node[right] {Time ($t$)};
    \draw[->] (0,0) -- (0,3) node[left] {VNE $S(\rho_t)$ (bits)};

    \draw (1,0) -- (1,-0.1) node[below] {$t=0$};
    \draw (2.5,0) -- (2.5,-0.1) node[below] {$t=1$};
    \draw (4,0) -- (4,-0.1) node[below] {$t=2$};
    \draw (0,1*\yscale) -- (-0.1,1*\yscale) node[left] {1};
    \draw (0,0) -- (-0.1,0) node[left] {0}; 

    \draw[thick, vnecolor] plot coordinates {(1, 0) (2.5, 1*\yscale) (4, 0)}; 
    \fill[vnecolor] (1,0) circle (3pt);
    \fill[vnecolor] (2.5,1*\yscale) circle (3pt);
    \fill[vnecolor] (4,0) circle (3pt);

    \node[align=center] at (1, 0.5) {Context\\(Pure)};
    \node[align=center] at (2.5, 1*\yscale + 0.5) {Tension\\($\rho_{\text{mix}}$)};
    \node[align=center] at (4, 0.5) {Resolution\\(Pure)};
\end{tikzpicture}
\caption{Conceptual evolution of Von Neumann Entropy (VNE) $S(\rho_t)$. The entropy peaks when ambiguity is maximal and represented as a mixed state ($\rho_{\text{mix}}$ at $t=1$).}
\label{fig:vne_dynamics}
\end{figure}
\subsubsection{The Role of Coherence: Beyond Classical Mixtures}

Crucially, the density matrix formalism captures phenomena beyond classical statistical mixtures. Consider the mixed state $\rho_1$ (denoted $\rho_{\text{mix}}$) from the previous example. It represents classical uncertainty.

Now consider a state of superposition, such as the pure state $\ket{\psi_+} = \frac{1}{\sqrt{2}}(\ket{A} + \ket{B})$. The corresponding density matrix is:
\begin{equation}
\rho_{\text{sup}} = \ket{\psi_+}\bra{\psi_+} = \begin{pmatrix} 0.5 & 0.5 \\ 0.5 & 0.5 \end{pmatrix}
\end{equation}
While the populations (diagonal elements) are identical to $\rho_{\text{mix}}$, the non-zero coherences (off-diagonal elements) indicate a stable phase relationship between $\ket{A}$ and $\ket{B}$. This state is pure ($\gamma(\rho_{\text{sup}})=1, S(\rho_{\text{sup}})=0$), representing a deterministic, yet superimposed, musical context.

\textbf{Musical Interpretation: Structural Ambiguity vs. Uncertainty.}
The distinction between $\rho_{\text{mix}}$ and $\rho_{\text{sup}}$ is crucial for modeling musical ambiguity. Consider a pivot chord used during modulation, which simultaneously functions in two different keys. Let $\ket{A}$ represent the interpretation in the original key and $\ket{B}$ in the target key.

\begin{itemize}
    \item \textbf{Classical Mixture ($\rho_{\text{mix}}$):} Represents epistemic uncertainty. The model (or listener) is unsure whether the context is A \textit{or} B. This state has high entropy ($S(\rho_{\text{mix}})=1$ bit), reflecting genuine confusion or lack of information.
    \item \textbf{Coherent Superposition ($\rho_{\text{sup}}$):} Represents structural ambiguity inherent in the composition. The pivot chord functions simultaneously as A \textit{and} B. This is a pure state ($S(\rho_{\text{sup}})=0$), reflecting the deterministic nature of the dual function. The coherences encode the stable relationship between these two interpretations.
\end{itemize}

The ability to represent and evolve these coherences distinguishes DM-RNN from classical probabilistic models. The evolution from $\rho_{\text{sup}}$ differs significantly from $\rho_{\text{mix}}$ because the dynamics (CPTP map) act upon the coherences, allowing the model to capture how structured musical relationships resolve based on learned interference patterns.

\subsection{Identifying Musical Eigenstates: Spectral Analysis of $\rho_t$}

Since $\rho_t$ is Hermitian, the Spectral Theorem guarantees its diagonalization.

\textbf{Eigen-Decomposition.}
\begin{equation}
\rho_t = U_t \Lambda_t U_t^\dagger = \sum_{i=1}^d \lambda_i(t) \ket{\psi_i(t)}\bra{\psi_i(t)}
\label{eq:eigen_decomposition}
\end{equation}
where $U_t$ is a unitary matrix whose columns are the eigenvectors $\ket{\psi_i(t)}$, and $\Lambda_t$ is the diagonal matrix of corresponding eigenvalues $\lambda_i(t)$.

\begin{itemize}
    \item \textbf{Eigenvectors ($\ket{\psi_i(t)}$):} These \textit{Musical Eigenstates} are the pure states constituting the ensemble $\rho_t$.
    \item \textbf{Eigenvalues ($\lambda_i(t)$):} They represent the probability distribution over the eigenstates.
\end{itemize}

\subsection{Analyzing Polyphonic Correlation: Entanglement}

The density matrix formalism allows for the rigorous modeling of correlations (entanglement) between subsystems.

\textbf{Bipartite System Representation.} Consider two voices, A and B. We analyze the system in a composite Hilbert space $\mathcal{H} = \mathcal{H}_A \otimes \mathcal{H}_B$. The joint state is $\rho_{AB} \in \D(\Hp)$.

\textbf{Prerequisite: Imposed Tensor Product Structure.} A critical prerequisite is that $\Hp_d$ must possess a well-defined tensor product structure ($\Hp_d \cong \Hp_A \otimes \Hp_B$) aligned with the musical voices, such that $d=d_A \times d_B$. This structure must be explicitly imposed architecturally.


\textbf{Architectural Enforcement of Structure.}
In the context of the DM-RNN, this imposition dictates the design of the parameterized Quantum Channel $\E_t$ (Section 4.4). The neural network $G$ must be structured to operate within this composite space.

A practical approach is to decompose the dynamics into local operations and interaction terms:

\begin{equation}
\E_t \approx (\E_A \otimes \E_B) \circ \E_{\text{int}}
\end{equation}
Here, $\E_A$ and $\E_B$ are local channels acting independently on $\Hp_A$ and $\Hp_B$, and $\E_{\text{int}}$ is an entangling channel modeling interactions. $G$ can be structured with dedicated modules generating the parameters for these component channels separately. This ensures that the learned hidden state $\rho_t$ (representing $\rho_{AB}$) maintains a meaningful factorization corresponding to the musical voices.

\textbf{Reduced Density Matrices.} Assuming an imposed structure, the state of voice A is obtained via the partial trace over subsystem B, $\Tr_B: \Lop(\Hp_A \otimes \Hp_B) \rightarrow \Lop(\Hp_A)$.

\begin{equation}
\rho_A = \Tr_B(\rho_{AB})
\label{eq:partial_trace}
\end{equation}
This operation effectively ``traces out'' the influence of subsystem B. Explicitly, if $\{\ket{j}_B\}$ is an orthonormal basis for $\Hp_B$, the partial trace is computed as:
\begin{equation}
\rho_A = \sum_j (\Id_A \otimes \bra{j}_B) \rho_{AB} (\Id_A \otimes \ket{j}_B)
\end{equation}
where $\Id_A$ is the identity operator on $\Hp_A$.

\textbf{Quantifying Correlation via Quantum Mutual Information.} We use Quantum Mutual Information (QMI)~\cite{Wilde2017}, $I(A;B)$, to measure the total correlation (both classical and quantum) between the subsystems:

\begin{equation}
I(A;B) = S(\rho_A) + S(\rho_B) - S(\rho_{AB})
\label{eq:qmi}
\end{equation}
$I(A;B) \ge 0$, with equality iff the systems are uncorrelated ($\rho_{AB} = \rho_A \otimes \rho_B$). For mixed states, while QMI provides an upper bound on the distillable entanglement, it serves as a robust measure of the overall interdependence between the voices.

\textbf{Example: Separable vs. Entangled States.}
To illustrate, consider a minimal bipartite system where $\Hp_A = \Hp_B = \C^2$.
\begin{enumerate}
    \item \textbf{Separable State (Uncorrelated):} If the voices are independent, the state is separable. Let $\rho_{AB}^{\text{sep}} = \rho_A \otimes \rho_B$. Suppose $\rho_A = \rho_B = \ket{0}\bra{0}$. Then $\rho_{AB}^{\text{sep}} = \ket{00}\bra{00}$. The entropies are $S(\rho_A)=S(\rho_B)=S(\rho_{AB}^{\text{sep}})=0$. Thus, $I(A;B) = 0$.

    \item \textbf{Maximally Entangled State (Correlated):} Consider the Bell state $\ket{\Psi^+} = \frac{1}{\sqrt{2}}(\ket{00} + \ket{11})$. The joint state $\rho_{AB}^{\text{ent}} = \ket{\Psi^+}\bra{\Psi^+}$ is pure, so $S(\rho_{AB}^{\text{ent}})=0$. However, the reduced density matrices (obtained via Eq.~\ref{eq:partial_trace}) are maximally mixed: $\rho_A = \Tr_B(\rho_{AB}^{\text{ent}}) = \frac{1}{2}\Id_2$. Thus, $S(\rho_A)=1$ bit (similarly $S(\rho_B)=1$). The QMI is $I(A;B) = 1+1-0 = 2$ bits. This signifies maximum correlation: the state of voice A perfectly predicts the state of voice B, despite the local state of A being maximally uncertain.
\end{enumerate}

\subsubsection{Musical Interpretation of QMI}

The dynamics of QMI provide a novel metric for analyzing musical texture and the interplay between voices.
\begin{itemize}
    \item \textbf{High QMI (Strong Correlation/Entanglement):} Expected during sections with high contrapuntal interdependence (e.g., fugues or canons), where the state of one voice strongly dictates the state of the other.
    \item \textbf{Low QMI (Weak Correlation/Separability):} Expected during homophonic textures, where voices move in rhythmic unison or one voice clearly dominates. The joint state approaches a separable state.
\end{itemize}
The temporal evolution of $I(A;B)$ can thus track the shifting dependency relationships between voices throughout a polyphonic composition.

\section{Discussion and Future Work}

The DM-RNN framework introduces a paradigm shift, moving beyond vector representations $h_t \in \R^d$ to embrace matrix-based states $\rho_t \in \D(\Hp_d)$.

\subsection{Theoretical Implications}

\textbf{Capacity for Expressing Ambiguity and Coherence.} By utilizing the density matrix, DM-RNN possesses the mathematical capacity to represent statistical ensembles (mixed states) and pure superpositions (coherence), natively capturing the inherent ambiguity of musical context.

\subsection{Computational Challenges}

\textbf{Scalability.} The DM-RNN scales the hidden state representation from $O(d)$ (for classical RNNs) to $O(d^2)$. More significantly, the update process introduces a substantial bottleneck in parameter complexity. As detailed in Section 4.4, parameterizing a general CPTP map requires the neural network $G$ to generate the Cholesky factor $L_t$, which inherently involves $O(d^4)$ parameters at each time step. For a modest Hilbert space dimension (e.g., $d=32$), this requires $O(32^4) \approx O(10^6)$ parameters per step, rendering the dense formulation practically intractable for complex musical structures.

\textbf{Enforcing Constraints.} Optimizing the model requires enforcing the TP constraint on the Choi matrix and the completeness constraint on the POVMs, which involve computationally intensive operations~\cite{Trouvain2020}.

\subsection{Implementation Strategies and Future Directions} 

\textbf{Tractable Implementation via Matrix Product Operators (MPO).} 
To address the prohibitive $O(d^4)$ parameter complexity of the dense parameterization, the practical implementation of the DM-RNN must leverage Tensor Network factorization~\cite{Orus2014}. We hypothesize that musically meaningful contexts often exhibit relatively low entanglement. In physics, many systems adhere to ``Area Laws'', where entanglement is limited~\cite{Eisert2010}. If musical contexts follow similar principles, the density matrix $\rho_t$ may be efficiently represented by Matrix Product States (MPS)~\cite{Schollwock2011} or, more accurately for mixed states, Matrix Product Density Operators (MPDO)~\cite{Verstraete2004}.

Crucially, the evolution operator (Quantum Channel $\E_t$) must be represented by Matrix Product Operators (MPO). MPOs provide an efficient representation by decomposing the high-dimensional evolution map (or its corresponding $O(d^4)$ parameters in $L_t$) into a network of lower-dimensional tensors, controlled by a parameter known as the \textit{bond dimension}, $\chi$. The bond dimension quantifies the entanglement capacity of the representation. If the dynamics can be accurately approximated with a small bond dimension ($\chi \ll d$), this dramatically reduces the parameter complexity from $O(d^4)$ to approximately $O(d^2 \chi^2)$. For the previous example ($d=32$), using a small bond dimension (e.g., $\chi=8$) reduces the complexity to $O(32^2 \cdot 8^2) \approx O(65,000)$, a significant improvement.
The implementation strategy involves defining the neural network $G$ (Section 4.4) such that it generates the dynamics in an MPO factored form. Utilizing MPOs is therefore essential to make the DM-RNN tractable.

\textbf{Architectures for Entanglement.} Developing specific DM-RNN architectures that impose and leverage the tensor product structure (Section 5.3) for polyphonic modeling is a critical next step.

\textbf{Theoretical Extensions: Open Systems.} Future research could explore modeling musical context within the framework of Open Quantum Systems. This approach naturally models musical ``dissipation'' or ``decoherence''—the gradual loss of coherence over time, perhaps representing the decay of harmonic tension or the forgetting of past context. When the evolution is assumed to be Markovian, the continuous-time dynamics are governed by the Lindblad master equation~\cite{Breuer2002}:
\begin{equation}
\frac{d\rho}{dt} = -i[H, \rho] + \sum_k \gamma_k \left(L_k \rho L_k^\dagger - \frac{1}{2}\{L_k^\dagger L_k, \rho\}\right)
\end{equation}
The first term describes coherent evolution via the Hamiltonian $H$, while the second term (the dissipator) describes decoherence via jump operators $L_k$ and rates $\gamma_k$, providing a rigorous framework for modeling the evolution and decay of musical expectations.

\section{Conclusion}

This paper introduced the Density Matrix Recurrent Neural Network (DM-RNN), a novel theoretical architecture utilizing the Density Matrix $\rho_t \in \D(\Hp_d)$ as the hidden state representation. This allows the model to natively capture the inherent ambiguity, coherence, and statistical mixture of expectations prevalent in music.

We have rigorously defined the architecture and its temporal dynamics using CPTP maps, providing a concrete mathematical construction for parameterizing these maps via the Choi-Jamio\l{}kowski isomorphism. The DM-RNN framework establishes a new paradigm for computational music analysis grounded in Matrix Theory. While computational challenges exist, the DM-RNN offers a mathematically rigorous approach to modeling complex musical structures.

%
%
\begin{credits}

\subsubsection{\discintname} 
The authors have no competing interests to declare that are relevant to the content of this article.
\end{credits}


\end{document}